\pdfoutput=1  

\documentclass[sigconf,natbib=false, nonacm]{acmart} 

\AtBeginDocument{%
  }

\RequirePackage[
  datamodel=acmdatamodel,
  style=acmnumeric,
  ]{biblatex}

\usepackage{algorithm}
\usepackage{algcompatible}
\usepackage{dashrule}

\algrenewcommand\algorithmicrequire{\textbf{Input:}}
\algrenewcommand\algorithmicensure{\textbf{Output:}}
\begin{document}

\title{BaKlaVa - Budgeted Allocation of KV cache for Long-context Inference}


\author{Ahmed Burak Gulhan}
\affiliation{%
  \institution{The Pennsylvania State University}
  \city{State College}
  \state{PA}
  \country{USA}
  }
\email{gulhan@psu.edu}

\author{Krishna Teja Chitty-Venkata}
\affiliation{%
  \institution{Argonne National Laboratory}
  \city{Lemont}
  \state{IL}
  \country{USA}
  }
\email{schittyvenkata@anl.gov}

\author{Murali Emani}
\affiliation{%
  \institution{Argonne National Laboratory}
  \city{Lemont}
  \state{IL}
  \country{USA}
  }
\email{memani@anl.gov}

\author{Mahmut Kandemir}
\affiliation{%
  \institution{The Pennsylvania State University}
  \city{State College}
  \state{PA}
  \country{USA}
  }
\email{mtk2@psu.edu}

\author{Venkatram Vishwanath}
\affiliation{%
  \institution{Argonne National Laboratory}
  \city{Lemont}
  \state{IL}
  \country{USA}
  }
\email{venkat@anl.gov}


\begin{abstract}
In Large Language Model (LLM) inference, Key-Value (KV) caches (KV-caches) are essential for reducing time complexity. 
However, they result in a linear increase in GPU memory as the context length grows. While recent work explores KV-cache eviction and compression policies to reduce memory usage, they often consider uniform KV-caches across all attention heads, leading to suboptimal performance. We introduce BaKlaVa, a method to allocate optimal memory for individual KV-caches across the model by estimating the importance of each KV-cache. Our empirical analysis demonstrates that not all KV-caches are equally critical for LLM performance. 
Using a one-time profiling approach, BaKlaVa assigns optimal memory budgets to each KV-cache. We evaluated our method on LLaMA-3-8B, and Qwen2.5-7B 
models, achieving up to a 70\% compression ratio while keeping baseline performance and delivering up to an order-of-magnitude accuracy improvement at higher compression levels. 
\end{abstract}




\maketitle

\section{Introduction}\label{sec:intro}

Large Language Models (LLMs) have achieved great success in recent years and have been successfully used in several natural language processing tasks such as chatbots, search engines, summarization, and customer service. This success has led to the development of LLMs with exponentially increasing parameter counts and context lengths (how much previous text an LLM can remember), with the latest models having more than a trillion parameters with more than a million context lengths~\cite{geminiteam2024, ren2023trillionparameter}. Although larger models with longer context lengths have improved performance, they come at the cost of significantly higher GPU memory usage during inference, posing challenges for efficient deployment.

LLMs generate text in an autoregressive manner -- given an input of any length, the model generates only a single token (word). To generate multiple tokens, the previously generated token is {\em appended} to the input, and the process repeats. This method of inference leads to significant `redundant computations', leading to quadratic time complexity. To mitigate this inefficiency, LLMs employ {\bf Key-Value (KV) caches} to store previous calculations -- key and value tokens for each attention head\footnote{An attention head is a key component of LLMs~\cite{vaswani2023attentionneed}, which allows the capture of the relationships between words. LLMs can have up to thousands of attention heads, each with their own KV-cache.} -- and remove these unnecessary computations. Yet, this comes at the cost of substantial GPU memory to hold these tokens, limiting how many tokens can be stored and, thus, a limit to how much an LLM can remember. This is currently one of the major bottlenecks in LLM scaling for long-context inference. 

Recent works have tried to address this challenge mainly by reducing the amount of data that an LLM needs to cache. However, many of these compression-based policies allocate memory {\em uniformly} across all KV caches, which is suboptimal. Recent research~\cite{squeezeattention, pyramidinfer, adakv}, has begun to explore the benefits of assigning heterogeneous memory budgets to different KV-caches. The key challenge in this approach lies in determining the optimal allocation of KV-cache memory, that is, which KV-caches in an LLM are more or less critical than the others to model performance.

In our work, {\bf BaKlaVa}, we demonstrate that different attention heads in an LLM have varying levels of importance, and therefore KV-cache memory should be allocated accordingly-- more important heads receiving larger (space for) KV-caches and less important ones receiving smaller allocations. To achieve this, we introduce a one-time `profiling' method, which does \textit{not} require fine-tuning of the LLM. Using a simple heuristic, our method estimates the importance of each attention head and optimally distributes a given KV-cache budget to maximize inference performance. We run multiple benchmarks with different KV-cache eviction and compression policies and show that our method can increase the inference quality up to an order of magnitude, without using additional memory or computation, and allows for near-baseline (a cache with maximum context length) performance for lower compression ratios.

BakLaVa is {\em complementary} to most existing KV-cache management and optimization methods, such as FlashAttention~\cite{flashattention2} and vLLM~\cite{pagedattention}, as well as various KV-cache compression, eviction, and offloading policies. We emphasize that our proposed method is {\em not} a policy for managing KV-cache memory or optimizing KV-cache calculations; rather, it is a method for \emph{allocating} memory budgets among existing KV-caches in an LLM.

\textbf{Contributions:} The main contributions of this paper can be summarized as follows:
\begin{itemize} 
    \item We introduce a heuristic to determine the relative importance of attention heads and KV caches in LLMs.
    \item We empirically show that the results of this heuristic remain consistent across various prompts for a given LLM. 
    \item We empirically validate that our heuristic {\em near optimally} ranks how important each KV cache is.
    \item We evaluate our proposed methodology on LongBench~\cite{bai2023longbench} and compare it against other KV-cache memory allocation strategies.
    \item Finally, we implement our method in HuggingFace as a custom KV-cache object.
\end{itemize}

Our empirical evaluations demonstrate that KV-cache and layer importance can be effectively estimated using heuristics with a high degree of accuracy. However, even with accurately identified importance values, determining the optimal memory allocation remains non-trivial, as the ideal strategy varies across different compression ratios. To address this challenge, we introduce a straightforward yet effective memory allocation approach that enables rapid parameter search, allowing us to efficiently determine the optimal memory distribution for various model architectures and compression ratios.
\section{Background}\label{sec:background}

\subsection{Self-Attention}

Consider an input matrix $\mathbf{Z} \in \mathbb{R}^{T \times D}$, where $T$ represents the sequence length and $D$ is the feature dimension. The multi-head self-attention mechanism facilitates learning from different representational subspaces by executing multiple attention computations in parallel. Query ($\mathbf{Q}$), Key ($\mathbf{K}$), and Value ($\mathbf{V}$) are derived from linear projections:
$\mathbf{Q} = \mathbf{Z} \mathbf{M}_Q$, $\mathbf{K} = \mathbf{Z} \mathbf{M}_K$, and $\mathbf{V} = \mathbf{Z} \mathbf{M}_V$, where $\mathbf{M}_Q, \mathbf{M}_K, \mathbf{M}_V \in \mathbb{R}^{D \times D_h}$ are trainable weight matrices. The attention weights are computed via scaled dot-product attention, as given in Eq. \ref{eq:attn_scaled}: 
\begin{equation}\label{eq:attn_scaled}
\text{Attention}(\mathbf{Q}, \mathbf{K}, \mathbf{V}) = \text{softmax}\left(\frac{\mathbf{Q} \mathbf{K}^\top}{\sqrt{D_h}}\right) \mathbf{V}.
\end{equation}
This process is repeated over $H$ heads, each utilizing distinct weight matrices $\mathbf{M}_Q^{(h)}, \mathbf{M}_K^{(h)}, \mathbf{M}_V^{(h)}$. The concatenated outputs from all heads are projected back to the original dimension $D$ using a learned weight matrix $\mathbf{M}_O \in \mathbb{R}^{H D_h \times D}$:
\begin{equation}
\text{MultiHead}(\mathbf{Q}, \mathbf{K}, \mathbf{V}) = \text{Concat}(\text{head}_1, \ldots, \text{head}_H) \mathbf{M}_O,
\end{equation}
where each attention head is defined as follows:
\begin{equation}
\text{head}_h = \text{Attention}(\mathbf{Q}^{(h)}, \mathbf{K}^{(h)}, \mathbf{V}^{(h)}).
\end{equation}

\subsection{Key-Value (KV) Cache} 
During autoregressive LLM inference, the tokens are generated sequentially. Without caching, Key ($\mathbf{K}$) and Value ($\mathbf{V}$) matrices are recomputed at each generation step for all preceding tokens. KV caching mitigates this inefficiency by storing computed $\mathbf{K}$ and $\mathbf{V}$ projections. Rather than recomputing these values, the model retrieves and appends the cached matrices to the current token’s projections. The updated attention computation follows Eq. \ref{eq:kv_cache_attn}: 
\begin{equation}\label{eq:kv_cache_attn}
\text{Attention}(\mathbf{Q}t, [\mathbf{K}{1:t-1}; \mathbf{K}t], [\mathbf{V}{1:t-1}; \mathbf{V}t]),
\end{equation}
where $[;]$ denotes concatenation along the sequence axis, and cached values ${\mathbf{K}{1:t-1}, \mathbf{V}_{1:t-1}}$ are loaded from memory. Although KV caching reduces redundant computation, storing cached projections for each token demands substantial memory, growing linearly with sequence length. For a transformer with $L$ layers, $H$ heads, and sequence length $T$, memory consumption scales as $2 \times T \times L \times H \times 16$-bit.

\subsection{KV-Cache Eviction} 
KV-Cache eviction aims to eliminate less significant tokens from $\mathbf{K}{1:t-1}$ and $\mathbf{V}{1:t-1}$ using a function $f_{evict}$ that identifies and removes redundant elements. The eviction mechanism is depicted in Eqs. \ref{eq:kv_evict_k} and \ref{eq:kv_evict_v}, where the $m^{th}$ token is removed from the cache.  
\begin{multline}\label{eq:kv_evict_k}
f_{evict}(\mathbf{K}{1:t-1}) = \mathbf{K'}{1:t-1} \\ = [k_1, \dots, k_{m-1}, k_{m+1}, \dots, k_{t-1}]
\end{multline}
\begin{multline}\label{eq:kv_evict_v}
f_{evict}(\mathbf{V}{1:t-1}) = \mathbf{V'}{1:t-1} \\ = [v_1, \dots, v_{m-1}, v_{m+1}, \dots, v_{t-1}]. 
\end{multline}
After eviction, attention is computed using the reduced cache, as shown in Eq.~\ref{eq:kv_attn_evict}: 
\begin{equation}\label{eq:kv_attn_evict}
\text{softmax}\left(\frac{\mathbf{Q}t [\mathbf{K'}{1:t-1}; \mathbf{K}t]^\top}{\sqrt{D_h}}\right) [\mathbf{V'}{1:t-1}; \mathbf{V}_t].
\end{equation}


\section{The BaKlaVa Method}\label{sec:methods}
Our method for optimizing KV-cache memory allocation consists of 3 main steps for a given LLM: (i) A one-time collection of profiling data for a given prompt(s) (Algorithm~\ref{alg:profiling} -- step 1);  (ii) Using a heuristic to estimate the `importance' of KV caches,  which is also a one-time calculation  (Algorithm~\ref{alg:profiling} -- step 2); and (iii) Performing a parameter search to allocate memory accordingly (Algorithms~\ref{alg:kv-mem-allocation} and~\ref{alg:parameter_search}). 

All three steps in BaKlaVa only need to be run once. The most time-consuming part currently is Step (iii), where a parameter search is performed for the target compression level. For this parameter search, we quickly evaluate each parameter combination using `perplexity', rather than running a long-context evaluation benchmark, since perplexity does not require autoregressive token generation and consequently is much faster and gives a good approximation of actual performance. This parameter search, for the models we evaluated (which contain 7 to 8 billion parameters), takes 10 to 20 minutes on 8x A100 GPUs for around 200 combinations of parameters on 98k tokens for a chosen compression ratio. The number of tokens can be decreased for a proportional decrease in runtime, though they should be at least as much as the maximum context length being evaluated.

Once the ideal parameters for an LLM are obtained, no additional computation is required.  To begin inference, we initialize our custom huggingface transformers' KV-cache object to use in inference.

\begin{algorithm}
\caption{One-Time Profiling for KV-Cache Importance}
\begin{algorithmic}[1]
\REQUIRE LLM model $\mathcal{M}$, one or more prompts $\mathcal{P}$ of varying lengths
\ENSURE Values indicating relative KV-cache importance

\STATE \textbf{Step 1:} Collect profiling data from prompts $\mathcal{P} = \{p_1, p_2, \dots, p_n\}$ of different lengths
\FOR{each prompt $p_i \in \mathcal{P}$}
    \STATE Run inference on $\mathcal{M}$ with $p_i$
    \FOR{each layer $l \in \mathcal{L}$}
        \FOR{each attention head $h$ in layer $l$}
            \STATE Compute token-wise cosine similarity between attention head input $V$ and output $SoftMax(QK^T) V$
            \STATE Compute the average across all cosine similarities to obtain a single similarity value $s_{il} \in \mathcal{S}$
        \ENDFOR
    \ENDFOR
\ENDFOR

\STATE \textbf{Step 2:} Convert attention head similarities to KV-cache importance using the number of attention heads per KV-cache group $g$
\FOR{each layer $l \in \mathcal{L}$}
    \FOR{each group of $g$ heads, denoted by $s_{il}, s_{(i+1)l}, \dots, s_{(i+g-1)l} \in \mathcal{S}$}
        \STATE Obtain similarity for the current KV cache, $KVsim \gets mean(s_{il}, \dots, s_{(i+g-1)l})$ 
        \STATE KV cache importance $I_{li} \gets 1-KVsim$
    \ENDFOR
\ENDFOR

\end{algorithmic}
\label{alg:profiling}
\end{algorithm}

\begin{figure}[h]
    \centering
    \vspace{-10pt}
    \includegraphics[width=0.8\linewidth]{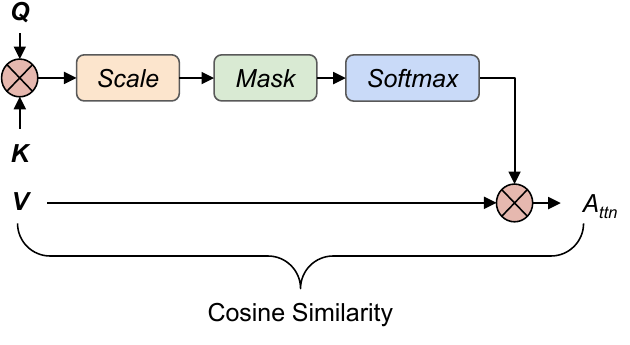}   
    \vspace{-0.25in}  
    \caption{The attention-head similarity heuristic used in BaKlaVa. By taking the cosine similarity between the input and output, we can calculate how much change there is. The more change between the input and output of the attention head, the more important we assume it is.}
    \label{fig:dot_product_cos} 
\end{figure}

\begin{figure}[h]
    \centering

    \includegraphics[width=1\linewidth]{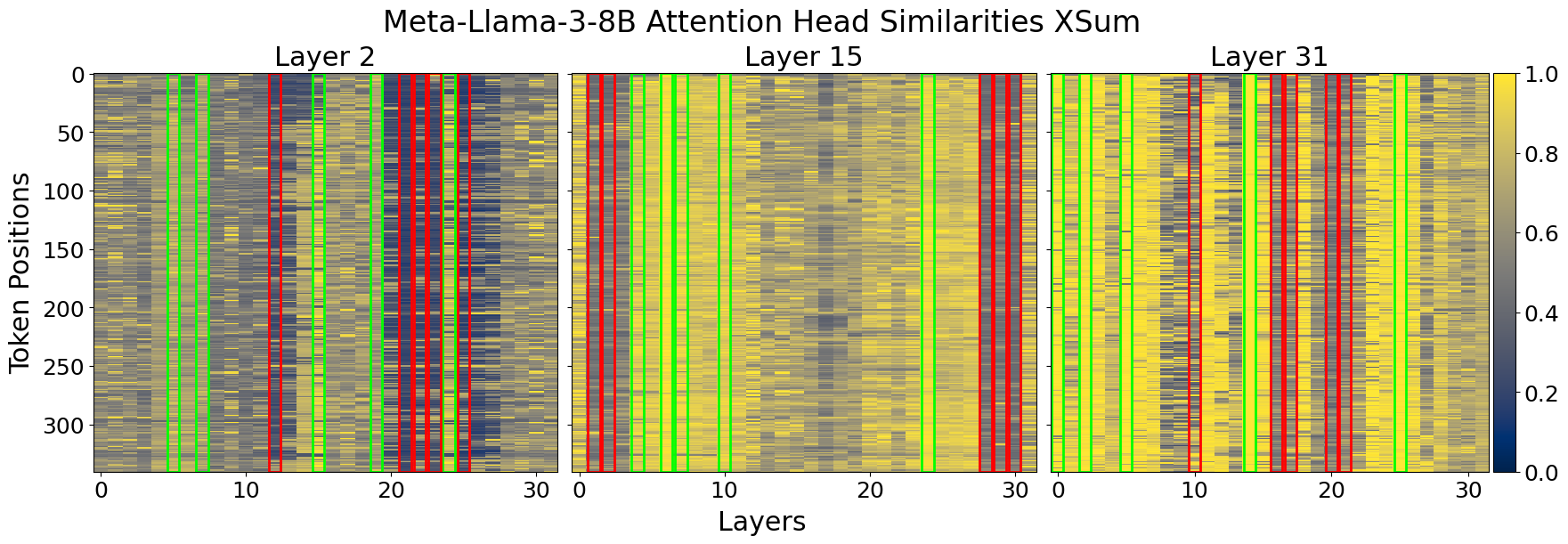}
    \includegraphics[width=1\linewidth]{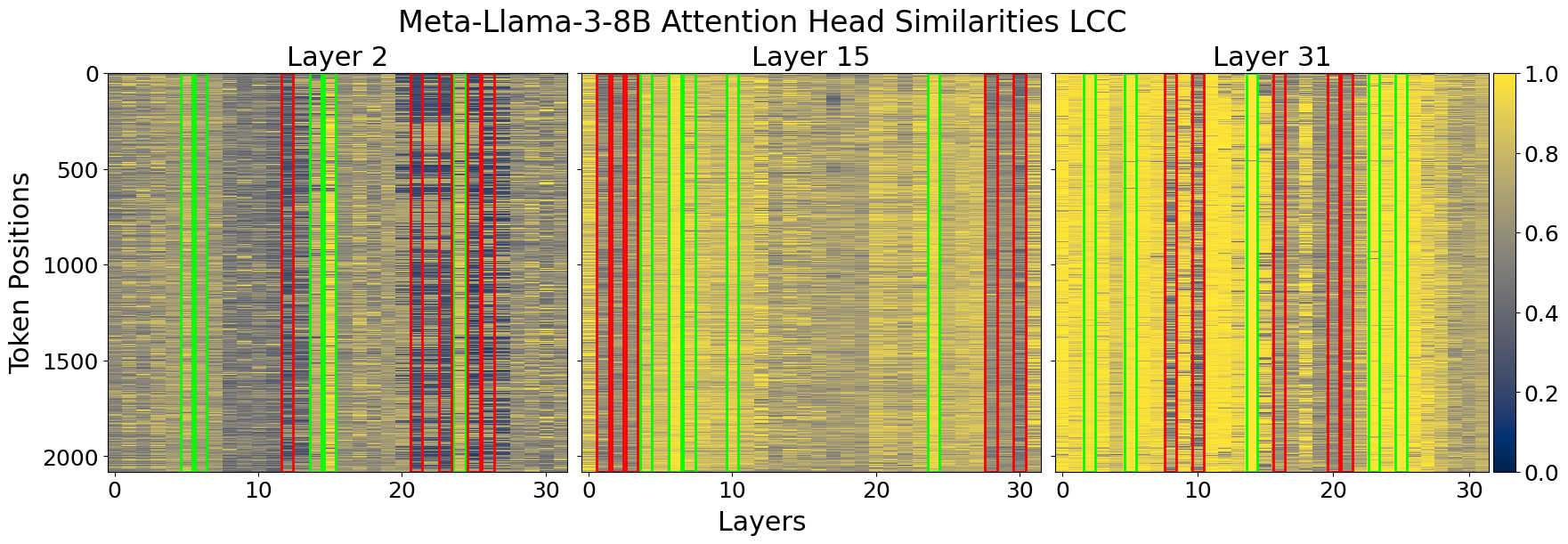}
    \hdashrule{\linewidth}{0.5pt}{1mm 1mm} 
    \includegraphics[width=1\linewidth]{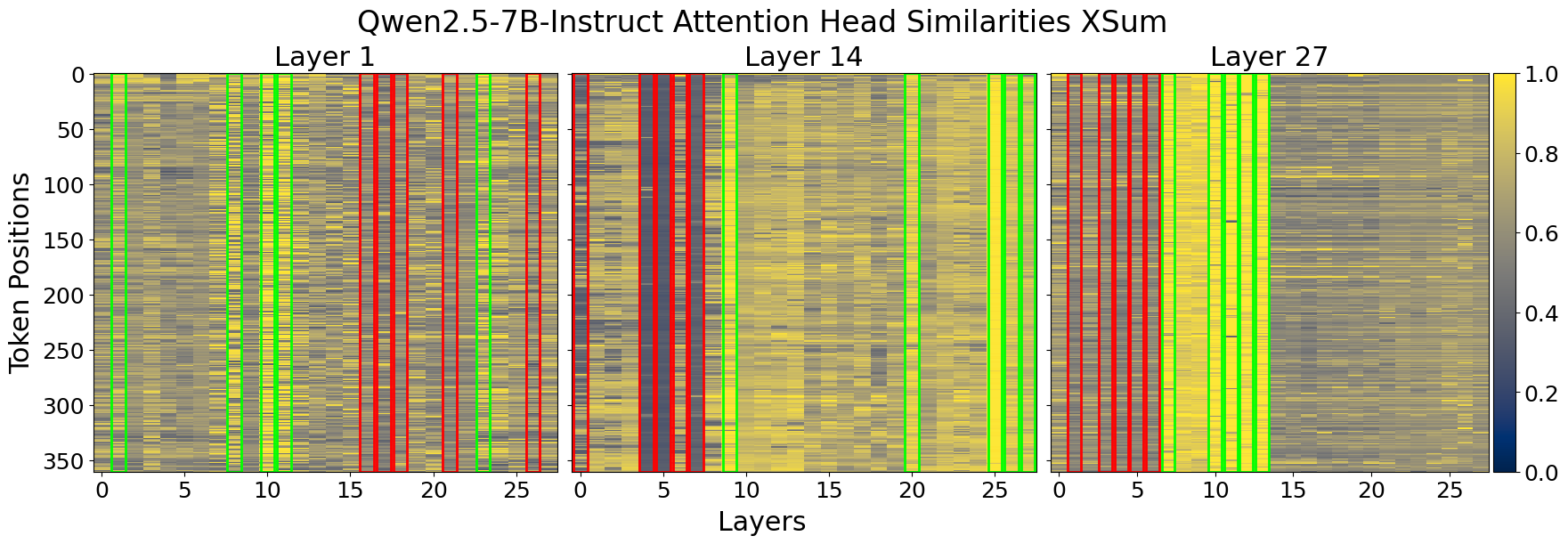}
    \includegraphics[width=1\linewidth]{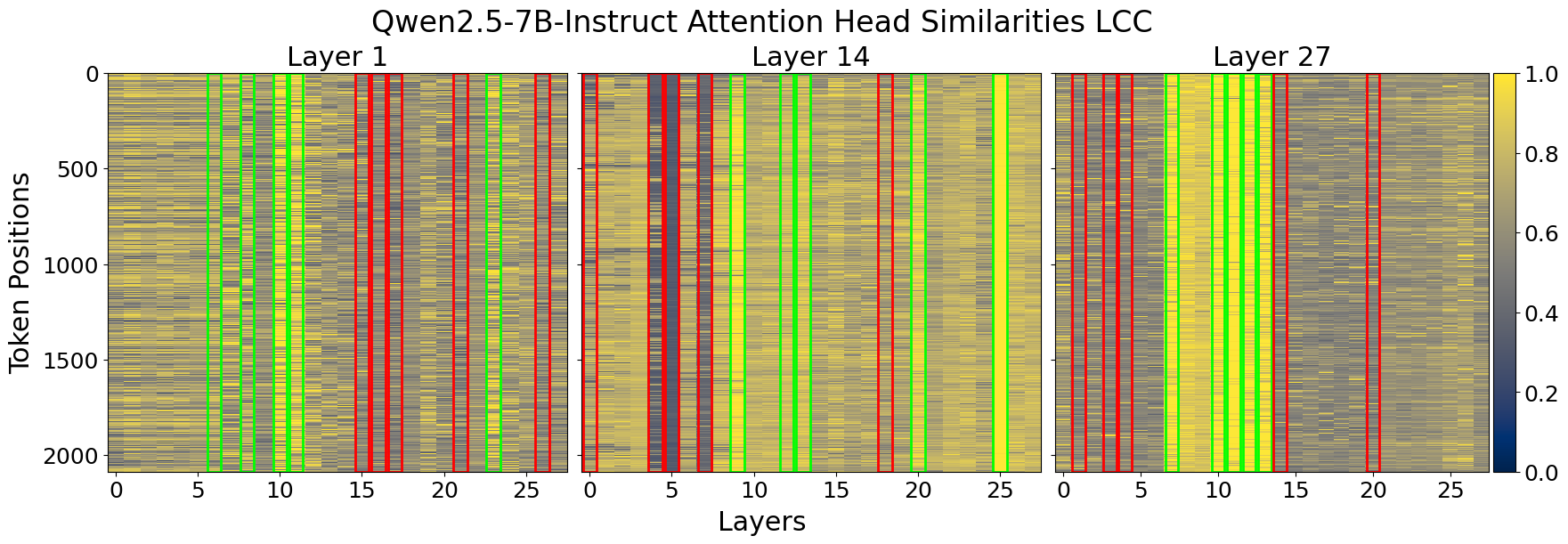}
    \caption{Cosine similarity heatmap for input and output of attention heads for two different prompts in LLaMA3-8B and Qwen2.5-7B. We chose three representative layers to illustrate that attention head consistency holds across different prompts. The X-axis shows the attention heads in a layer, Y-axis represents each token position in the prompt. Green and red outlines show the highest and lowest column similarity means per layer, that is, the most and least important attention heads respectively.  
    The order of average attention head similarities (the mean of each column, see Algorithm~\ref{alg:profiling}) stays highly consistent even across different prompts of different lengths, indicating that profiling an LLM one time is sufficient to make KV-cache importance estimations.}
    \label{fig:prompt_test} 
\end{figure}

\subsection{Determining KV-cache Importances}
\label{sec:determine_kv_importance}
\subsubsection{Head Importance Heuristic} 
To determine the significance of an individual attention head, we used several key observations to come up with a heuristic. The first is that the more change there is between the input and output of a structure in an LLM, the more important it is, as used in~\cite{ge2024model} to determine the type of tokens the individual attention heads focus on and in~\cite{squeezeattention, pyramidinfer} to determine the importance of LLM layers. Second, the attention matrix $softmax(QK^T)$ has been shown to be a high-quality heuristic that can determine individual token importances~\cite{h2o, scissorhands, keyformer} for KV-cache eviction. Lastly, the $V$ tensor contains key information about the tokens, which is not found in the attention score matrix~\cite{guo2024attention, devoto2024simple}. \textbf{Based on these observations, we propose the idea that the greater the change between the input} $V$ \textbf{and the output} $softmax(QK^T)V$ \textbf{ of this attention head, the more critical this attention head is for inference.}

As shown in Algorithm~\ref{alg:profiling}, to do this, we compare the input $V$ tokens (each token is a multidimensional vector) with the output of the attention head, $softmax(QK^T)V$, using cosine similarity, as shown in Figure~\ref{fig:dot_product_cos}. 
  
Each input and output token are individually compared using a cosine similarity value to determine the change in vector direction. To obtain a single value for each attention head, we first (i) get the mean of all tokens' cosine similarities within that head. The result of cosine similarity ranges from -1 to 1 and the maximum value is obtained when the two compared token vectors are identical (i.e., the attention score matrix is an identity matrix). We then (ii) normalize these values from range 0 to 1 to obtain a single similarity value, such that a value of 1 means identical input and outputs for all tokens in the attention head. Lastly, (iii) to obtain the `importance value', we take the complement of each mean similarity ($1 - similarity$). An importance value closer to 1 means a bigger difference between the input and output tokens, thus it has more importance and vice versa. 

We tested multiple token comparison methods, such as dot-product and KL divergence, and found that cosine similarity and dot-product both give similar results.  However, cosine similarity guarantees an output between -1 and 1, leading to simpler calculations. Therefore, we chose cosine-similarity for BaKlaVa. Note that cosine similarity measures the change in angle, but not magnitude, and thus another method that incorporates {\em both} magnitude and direction change may give better results. We left this for future work. 

\subsubsection{One-Time Profiling}
To determine the frequency of profiling needed, considering that attention head behavior can vary between inference steps and different prompts, 
we ran several experiments. We found that while the importance of individual tokens may change throughout inference, overall the importance value (calculated by taking the average of all token cosine similarities, see Algorithm~\ref{alg:profiling}, step 1) remains consistent, across different prompts. This is illustrated in Figure~\ref{fig:prompt_test}, where for two models, LlaMA3-8B and Qwen2.5-7B-Instruct, we profile two prompts of lengths around 350 and 2000 tokens from a text (XSum) and coding (LCC) dataset respectively, for three layers in both LLMs (not just to highlight these specific layers, but to illustrate that the consistency the attention head behavior holds across different layers). Each tile (in the heatmaps) represents the cosine similarity difference for a single token with the top 5 and bottom 5 importance heads outlined in green and red,  respectively. We can observe that the highest- and lowest-ranking attention heads stay highly consistent across prompts of different lengths and different types (i.e. text vs code). This suggests that, for a given LLM, a single profiling run with a sufficiently large prompt (i.e. few hundred tokens) is sufficient to determine the attention head importance values that can be applied for all future inferences.

\subsubsection{Grouped Query Attention} 
If the LLM employs Grouped Query Attention (GQA), an additional step is required before determining the KV cache memory budgets. Since our measurements assess changes at the level of individual attention heads rather than KV-caches, a direct assignment is not possible. In GQA, multiple attention heads \textit{share} the same KV-cache, meaning that memory budgets cannot be allocated separately for each head within the same group. To address this, we compute the mean of similarities across all attention heads within a group, obtaining a single similarity value per GQA group. This process is detailed in Step 2 of Algorithm~\ref{alg:profiling}

\subsubsection{Layer Importance Heuristic}
\label{sec:layer_heuristic}
The KV-cache importances we have found so far are used to allocate the GPU memory budget {\em within} a single layer in an LLM. Simply taking the average of all KV-cache importances does not find the correct layer importance, since our KV-cache importance heuristic is agnostic to several other important structures in an LLM (e.g., the feed-forward networks, layer normalization, etc). Based on our empirical testing results (see Section~\ref{sec:empirical}), we found that SqueezeAttention~\cite{squeezeattention} is a simple and low-overhead heuristic that closely, though not perfectly, matches the `ground truth' layer importances and use this heuristic to determine the layer-wise importance values in BaKlaVa. The SqueezeAttention heuristic takes the cosine similarity between the input and output of each LLM layer, thus capturing the total effect of {\em all} structures within the layer.

\subsection{Assigning Memory Budgets to KV-Caches}

\subsubsection{Memory Allocation}
\label{sec:methods-memory-alloc} 

\begin{algorithm}
\caption{KV-Cache Memory Reallocation Based on Attention Head Importance}
\begin{algorithmic}[1]
\REQUIRE Importance scores $\mathbf{I} = \{I_1, I_2, ..., I_m\}$ for $m$ KV-caches, threshold $t$, reduction amount $r$
\ENSURE Adjusted KV-cache allocations

\STATE $\mathcal{L} \gets \{i \mid I_i < t\}$ \COMMENT{Identify KV-caches with low importance}

\IF{$|\mathcal{L}| > m - 1$} 
    \STATE RETURN UNCHANGED KV-cache allocations 
    \COMMENT{If all KV-caches are low importance then do not do anything}
\ENDIF

\FOR{EACH $i \in \mathcal{L}$}
    \STATE REDUCE KV-cache size of $i$ by $r$
\ENDFOR

\STATE $n \gets |\mathcal{L}|$ \COMMENT{Number of KV-caches reduced}
\STATE $k \gets \min(n, m - n)$ \COMMENT{Limit reallocation up to the top n available high-importance caches}
\STATE $\mathcal{H} \gets$ TOP-$k$ ELEMENTS OF $\mathcal{H}$ BASED ON $I_i$ 
\STATE $\Delta r \gets \frac{n \times r}{k}$ \COMMENT{Compute adjusted increase per cache}

\FOR{EACH $j \in \mathcal{H}'$}
    \STATE INCREASE KV-cache size of $j$ BY $\Delta r$
\ENDFOR

\STATE RETURN updated KV-cache allocations

\end{algorithmic}
\label{alg:kv-mem-allocation}
\end{algorithm}

Once the importance values for each KV-cache and layer are obtained, the next step is to determine how to allocate memory budgets. 

Based on our observation of token similarities as shown in Figure~\ref{fig:prompt_test}, we find that low-importance attention heads are more consistent with how they change each individual token in a prompt (that is, the dot-product between input and output of the attention head has low variance), while other attention heads can display significant changes across tokens (that is, high variance in the token cosine similarity). Thus, to reduce the chances for decreasing the memory of KV-caches belonging to potentially critical attention heads, we take a conservative approach and only target KV-caches with an importance score below a threshold $t$ by a predetermined amount $r$, as shown in Algorithm~\ref{alg:kv-mem-allocation}. The freed memory is then assigned to up to top $n$ KV caches of highest importance (where $n$ is the number of low-importance KV-caches selected), in order to prioritize increasing memory for the most important KV-caches.

\subsubsection{Parameter Search}
\label{sec:parameter_search}

\begin{algorithm}
\caption{Parameter Search Using Perplexity}
\begin{algorithmic}[1]
\REQUIRE Evaluation prompt $\mathcal{P}$, model $M$, context length $L$, list of parameter configurations $\mathcal{C}$, compression ratio $cmp$
\ENSURE Optimal parameters $p^*$

\STATE $best\_params \gets \emptyset$
\STATE $min\_loss \gets \infty$

\FOR {$\text{params} \in \mathcal{C}$}
    
    \STATE CACHE = $\text{MAKE\_CACHE}(\text{params}, cmp)$
    \STATE $losses \gets []$
    
    \FOR{$\text{chunk} \in \text{STEP}(\mathcal{P}, L)$} \COMMENT{Get chunks of tokens from prompt}
        \STATE $loss \gets \text{PERPLEXITY}(M, \text{chunk}, CACHE)$
        \STATE $losses.\text{append}(loss)$
        \STATE CACHE $\gets \text{RESET\_CACHE}(CACHE)$
    \ENDFOR

    \STATE $avg\_loss \gets \frac{\sum \text{losses}}{\text{len}(\text{losses})}$

    \IF{$avg\_loss < min\_loss$}
        \STATE $min\_loss \gets avg\_loss$
        \STATE $best\_params \gets \text{params}$
    \ENDIF

\ENDFOR

\STATE RETURN $best\_params$

\end{algorithmic}
\label{alg:parameter_search}

\end{algorithm}

To determine the optimal values for $r$ and $t$, we performed a parameter search over different compression values, as shown in Algorithm~\ref{alg:parameter_search}. We found that the ideal parameter `area' varies across different compressions. These results were calculated using perplexity, as it is much faster to find compared to LongBench and it is a good indicator of actual performance. Based on these observations, we chose the best-performing parameter pair for a given compression value when evaluating on LongBench. 

\subsection{Empirical Evaluation of Heuristics}
\label{sec:empirical}

To evaluate the effectiveness of our layer and KV-cache importance heuristics in comparison to the `true' importance, we conducted computationally intensive experiments. These tests empirically assessed the impact of individual layers
on model performance by measuring the variation in benchmark scores resulting from modifications to each component. The underlying principle is that the greater the performance degradation caused by a change (e.g., memory reduction) in a layer or KV-cache, the more critical that component is to the model’s overall functionality.


To evaluate the layer importance heuristic (see Section~\ref{sec:layer_heuristic})  we tested our LLM on the \textit{triviaqa} LongBench dataset after reducing the memory allocated to different groups of layers. We selected a single dataset to minimize the computational cost of our empirical evaluation. \textit{triviaqa} was specifically chosen because it exhibits the widest range of scores, making it more sensitive to performance variations and thus a better candidate for detecting changes in output.

As described in algorithm~\ref{alg:layer_empirical}, we systematically reduced the memory budgets of layers within a sliding window of size 5, running a separate benchmark for each window position. Rather than evaluating individual layers in isolation, we compressed groups of 5 adjacent layers at a time. If crucial layers were arbitrarily scattered, rather than forming coherent clusters, it would suggest an unintuitive and unlikely distribution of importance. Additionally, testing each layer in isolation (i.e., using a window size of 1) yielded erratic results, indicating that individual layer evaluations do not capture meaningful patterns of layer importance. By considering contiguous groups, we aim to better approximate the true structure of importance within the model.

\begin{algorithm}
\caption{Empirical Evaluation of Layer Heuristic}
\begin{algorithmic}[1]
\REQUIRE LLM $M$, window size $W$, benchmark $BENCH$
\ENSURE Scores for each layer $S$, compression ratio $cmp$

\STATE $S \gets [\ ]$ \COMMENT{Initialize empty list for scores}

\FOR {$L \in \{0, \dots, \text{last\_layer}(M)\}$}
    \STATE $L_{\text{min}} \gets \max(0, L - \lfloor W/2 \rfloor)$
    \STATE $L_{\text{max}} \gets \min(\text{last\_layer}(M), L + \lfloor W/2 \rfloor)$

    \STATE $\text{reduce\_kv\_cache}(M, L_{\text{min}}, L_{\text{max}}, cmp)$ \COMMENT{Reduce KV-cache sizes for layers in window}

    \STATE $\text{score} \gets BENCH(M)$ \COMMENT{Run LLM and obtain benchmark score}

    \STATE $S.\text{append}(\text{score})$
\ENDFOR

\STATE RETURN $S$

\end{algorithmic}
\label{alg:layer_empirical}
\end{algorithm}

\section{Results}\label{sec:results} 
In this section, we report the results of BaKlaVa for KV-cache compression on the models LLaMA-3-8B and Qwen2.5-7B. 
Qwen weights are quantized to 8 bits due to hardware limitations. Section~\ref{sec:longbench_results} shows the results of LongBench on different KV-cache reduction methods for different compression ratios. Section~\ref{sec:empirical_results} reports the results of empirically evaluating how well the heuristics used in BaKlaVa reflect actual layer and KV-cache importances.

\subsection{LongBench}
\label{sec:longbench_results}

\begin{figure*}[hbt!]
    \centering
    
    \includegraphics[width=\textwidth]{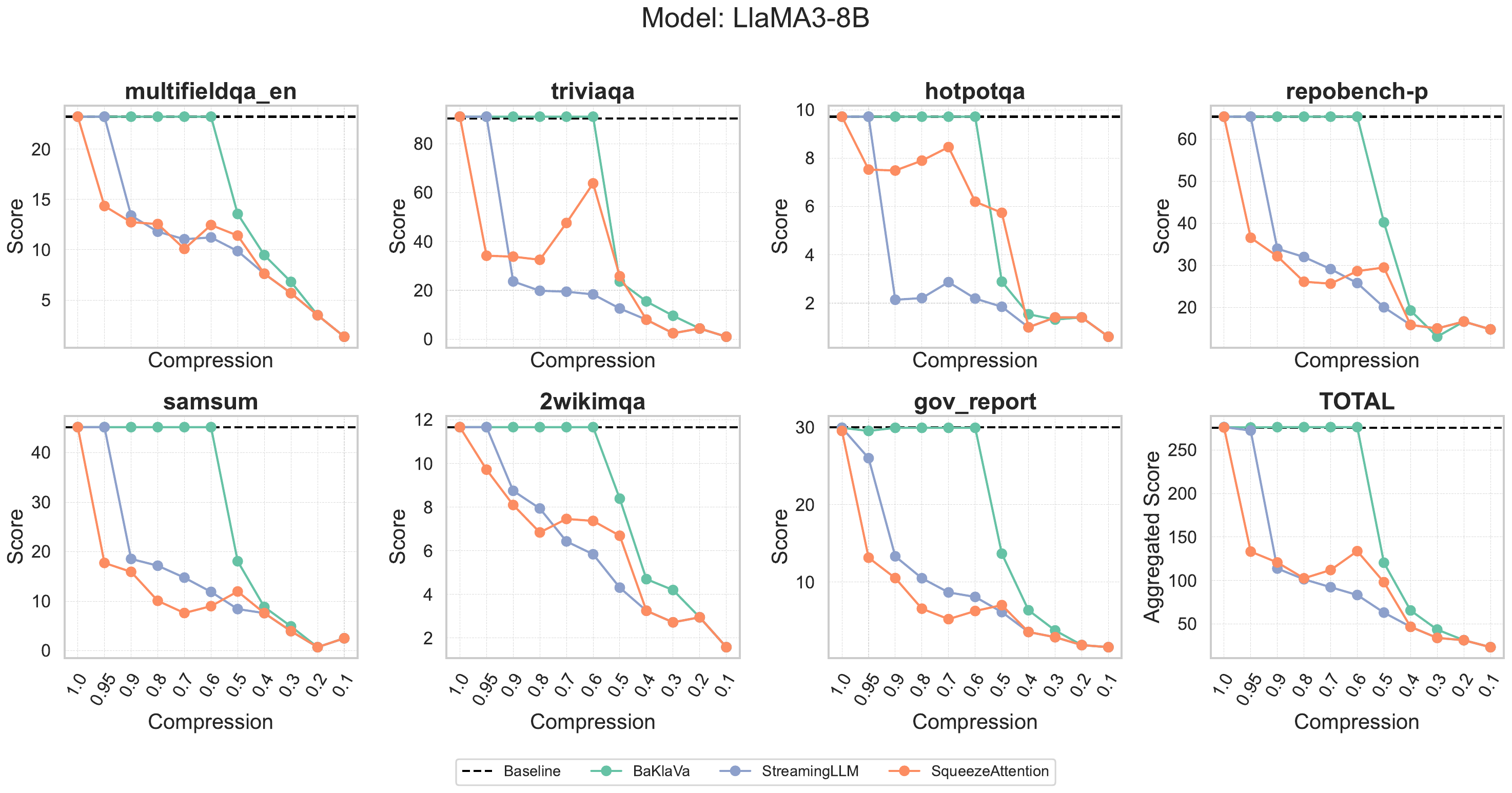}
    \includegraphics[width=\textwidth]{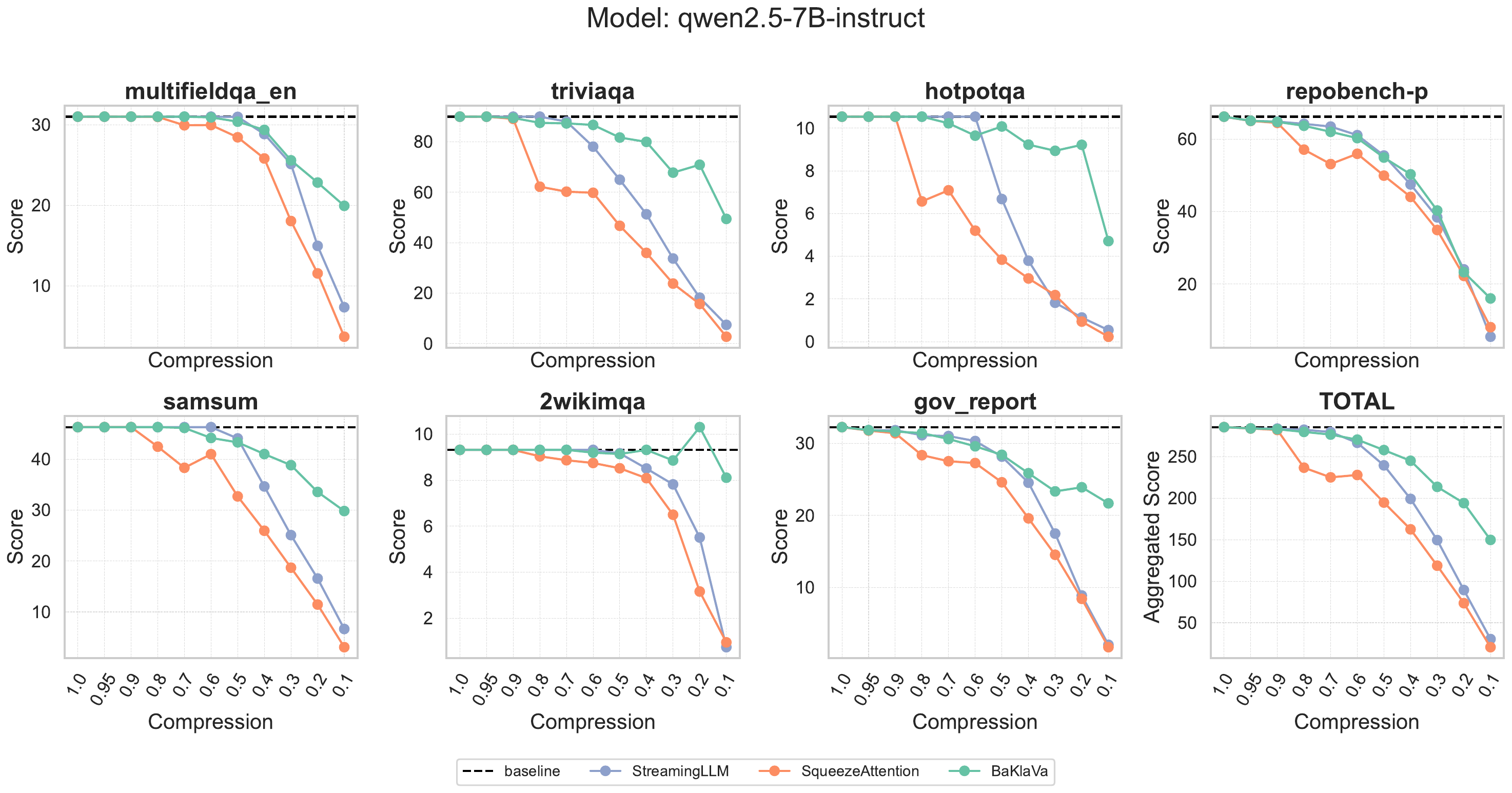}

    \caption{Comparison of BaKlaVa (varying memory budgets for both layers and KV-caches), SqueezeAttention (varying memory budgets for layers), and StreamingLLM (uniform memory budget for all KV-caches) on different LongBench tasks under various compression settings. The LongBench datasets shown include few-shot learning(triviaqa), coding (repobench-p), multi-document question answering (2wikimqa), and summarization (gov\_report). For BaKlaVa and SqueezeAttention, we conducted a parameter search using perplexity as a benchmark to determine the optimal settings for each compression ratio (detailed in Section~\ref{sec:parameter_search}).}

    \label{fig:longbench_results_small}
\end{figure*}

We used the LongBench~\cite{bai2023longbench} evaluation suite to test how our proposed approach works with real-life scenarios with significant KV-cache memory usage. LongBench contains 14 English (\textit{qmsum, multifieldqa\_en, triviaqa, hotpotqa, samsum, musique, multi\_news, 2wikimqa, gov\_report, trec,narrativeqa, passage\_count, passage \_retrieval\_en,} and \textit{qasper}) and 2 coding tasks (\textit{lcc} and \textit{repobench-p}), with average contexts for these tasks ranging between 5000 and 15000 tokens -- though the longest contexts exceed 32000 tokens.

Note that LongBench's default prediction script allows truncating prompts longer than a user-defined threshold (the $max\_length$ parameter) before being input into the LLM model, so that the instructions at the beginning or end of the prompt are not removed. We used the default LongBench configuration, where the prompt is truncated to a size below the maximum context length of the LLM. For our results, this equals a $max\_length$ of 7500 tokens if the context length is 8196 (LlaMA-3 8B) and 31500 tokens for a context length of 32768 (qwen-2.5 7B)

LongBench results for \textit{triviaqa, samsum} (few-shot learning), \textit{repobench-p} (coding), \textit{2wikimqa, hotpotqa} (multi-document Q\&A), \textit{multifieldqa\_en} (single-document Q\&A), \textit{gov\_report} (summarization) and the total aggregate scores of all 16 English and coding tasks are plotted in Figure~\ref{fig:longbench_results_small}. Higher values indicate better performance.  
\\
We evaluate three KV-cache eviction and memory allocation strategies: StreamingLLM~\cite{streamingllm}, SqueezeAttention~\cite{squeezeattention}, and BaKlaVa. Strea-mingLLM applies a token eviction policy once a KV-cache reaches capacity but assigns equal memory to all caches. SqueezeAttention extends this approach by varying memory allocation across layers, while BaKlaVa further refines memory distribution by adjusting allocations for both layers and individual KV-caches. Note that other token eviction or compression policies can be used with SqueezeAttention and BaKlaVa; we chose StreamingLLM for ease of implementation. 

In Figure~\ref{fig:longbench_results_small} the total aggregate score shows that BaKlaVa outperforms other KV-cache memory allocation methods (StreamingLLM and SqueezeAttention) on average; however, the results show distinctly different behaviors for LlaMA3-8B and Qwen2.5-7B models. This is due to the different context lengths of both models (8196 tokens for LlaMA3-8B and 32768 for Qwen2.5-7B) along with the prompt truncation behavior of LongBench we discussed above. In Qwen2.5-7B, at compression ratio 1.0 we start with the default context length of 32k tokens, which is significantly more than most prompts in LongBench. Thus, increasing KV-cache compression does not affect results up to a certain point (e.g., 0.5 compression ratio in \textit{multifieldqa\_en} for Qwen). Afterward, there is a gradual decrease in the score as long-context prompts lose critical information upon compression. 

On the other hand, for LlaMA3, at compression ratio 1.0 we start at the model's maximum context length of 8192 tokens. Since most prompts in LongBench are more than 8192 tokens, they get truncated in the middle, leaving the critical instructions at the beginning and/or end of the now shorter prompt. Thus, in LlaMA, the score starts decreasing rapidly upon increasing the KV compression, as the critical instructions at the beginning of the prompt get quickly evicted with the StreamingLLM window-based eviction policy. With BaKlaVa, we observe that these critical instructions are remembered until a much higher compression ratio. For example, 0.6 compression for all results in Figure~\ref{fig:longbench_results_small} for LlaMA. This behavior is also observed in Qwen, but for higher compression rations of 0.1 and 0.2, where BaKlaVa retains an order of magnitude larger score compared to the other methods. This can be seen in \textit{multifieldqa\_en, samsum, 2wikimqa}, etc. for Qwen. 

Finally, we see that on average SqueezeAttention performs better than StreamingLLM in LlaMA, but worse in Qwen. This is due to the layer importance heuristic used in SqueezeAttention closely matching the `true' importance values in LlaMA, but not in Qwen. See Section~\ref{sec:empirical_results} for details.

\subsection{Empirical Evaluation of Heuristics}
\label{sec:empirical_results}

\begin{figure}
    \centering
    \includegraphics[width=1\linewidth]{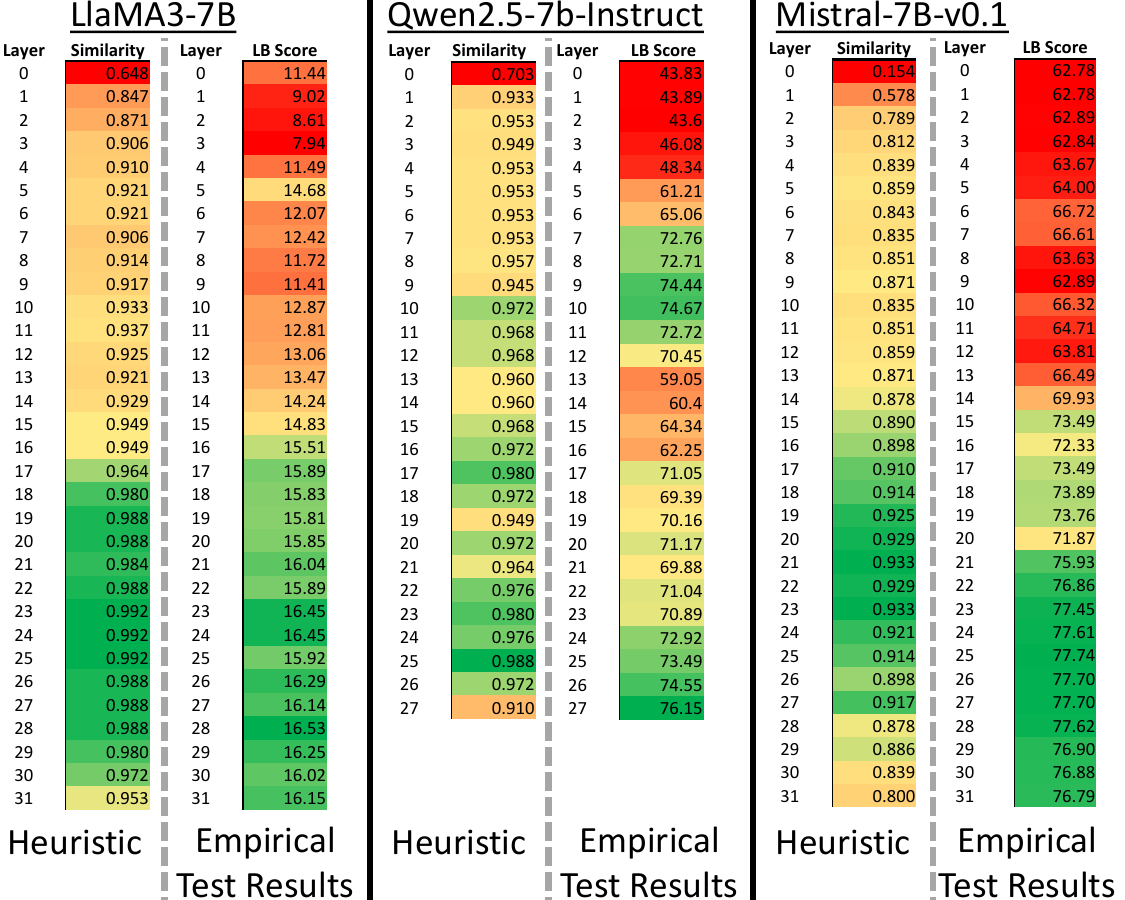}
    \caption{Comparison of layer importance heuristics with empirical evaluation results. Layers identified as least important by both heuristic and empirical test scores are highlighted in green, while critical layers are marked in red. LlaMA3-8B and Mistral-7B-v0.1 exhibit strong alignment between heuristic predictions and empirical findings, whereas Qwen2.5-7B shows significant discrepancies. Consequently, applying layer-wise KV-cache memory allocation based on the heuristic to Qwen2.5-7B may result in performance degradation.}
    \label{fig:empirical_layer_results}
\end{figure}

Figure~\ref{fig:empirical_layer_results} presents the empirical test results (detailed in Section~\ref{sec:empirical}) for the LlaMA3-8B, Qwen2.5-7B, and Mistral-7B models, compared against the layer importance heuristics used in BaKlaVa and SqueezeAttention for estimation of layer importance with low computational cost. The heuristic results are expressed as similarity scores, as defined in Section~\ref{sec:determine_kv_importance}, while the empirical results are reported as average LongBench scores on the \textit{triviaqa} dataset across compression ratios ranging from 0.5 to 0.95. Both heuristic and empirical results are visualized using a red-yellow-green color gradient, where less critical layers (i.e., those with high similarity scores or high empirical performance) are highlighted in green. It is important to note that the first and last two empirical test results exhibit a bias toward higher scores due to the sliding window approach. This overlap at the layer boundaries results in less cumulative compression across layers, thereby preserving more KV-cache memory overall.

Accounting for this bias, we observe that LlaMA3-8B and Mistral-7B exhibit highly similar layer importance patterns, as indicated by the alignment of green and red-shaded regions. In contrast, Qwen2.5-7B demonstrates a markedly different layer importance distribution. Consequently, applying layer-wise KV-cache memory allocation to Qwen leads to poorer performance compared to equal memory allocation, as evidenced by the Qwen results for SqueezeAttention and StreamingLLM in Figure~\ref{fig:longbench_results_small}.

\section{Related Works}\label{sec:related}
In this section, we discuss previous work relevant to BaKlaVa in five main areas: KV-cache eviction policy, profiling for determining memory budget, KV-cache quantization, cache merge, and system-level optimizations. 

\subsection{KV Cache Eviction Policy}
StreamingLLM~\cite{streamingllm} discovered the 'attention sink' effect, where early sequence tokens play a crucial role in maintaining model performance through asymmetric attention weight accumulation. H2O~\cite{h2o} introduces an eviction strategy based on cumulative attention, retaining 'heavy-hitter' key-value pairs while allowing token positions to vary. Similarly, Scissorhands~\cite{scissorhands} develops an approach that evicts based on a 'pivotal' metric, adjusting eviction rates across layers using a persistence ratio. Keyformer~\cite{keyformer} addresses the issue of token removal distorting softmax probability distributions by implementing regularization techniques to mitigate these perturbations. 

\subsection{Profiling for Determining Memory Budget}
Squeezeattention~\cite{squeezeattention} employs a dynamic approach, measuring layer importance through cosine similarity of input prompt differences pre- and post-self-attention, subsequently categorizing layers and adjusting their KV budgets. PyramidInfer~\cite{pyramidinfer} introduces a pyramid-shaped allocation strategy, prioritizing tokens with high attention values and maintaining a set of significant tokens through attention-driven updates during decoding. In comparison, Ada-KV~\cite{adakv} offers an adaptive budget allocation method that improves utilization across individual attention heads, resulting in more effective cache eviction strategies.

\subsection{KV-Cache Quantization}
GEAR~\cite{gear} takes a different approach by compressing less important entries to ultra-low precision, using a low-rank matrix for residual error approximation, and utilizing a sparse matrix for outlier correction. MiKV~\cite{notokenleftbehind} introduces a mixed-precision KV-cache quantization method, allocating precision based on token importance. QAQ~\cite{qaq} proposes a dynamic, quality-adaptive quantization approach that determines bit allocation based on token importance and sensitivity. KVQuant~\cite{hooper2024kvquant} offers strategies for smooth quantization of keys and values, including pre-RoPE quantization for keys, per-token quantization for values, and isolation of outliers in a sparse format. These diverse techniques collectively contribute to significant improvements in model compression and efficiency while maintaining performance.

\subsection{Cache Merge}
MiniCache~\cite{minicache} leverages the high angular similarity observed in middle-to-deep layer KV caches, merging key and value pairs from adjacent similar layers into shared representations. KVSharer~\cite{yang2024kvsharer}, on the other hand, exploits the counterintuitive finding that sharing KV caches between significantly different layers during inference does not substantially impact performance, prioritizing dissimilar layers for sharing based on Euclidean distance calculations. In comparison,  CaM~\cite{cam} focuses on merging keys or values of multiple evicted tokens with retained tokens using attention scores, while KVMerger~\cite{wang2024model} employs a two-step process: first clustering consecutive tokens with high cosine similarity, then merging tokens within each set into a pivotal token chosen by the attention score, using Gaussian kernel weights to emphasize contextual relevance. 

\subsection{System-Level Optimizations}
FlexGen~\cite{flexgen} proposes an SSD-based method for managing key-value (KV) caches, effectively expanding the memory hierarchy across GPU, CPU, and disk storage. This approach utilizes linear programming to optimize tensor storage and access patterns, enabling high-throughput LLM inference on hardware with limited resources. Complementing this, ALISA~\cite{alisa} introduces a dual-level KV cache scheduling framework that combines algorithmic sparsity with system-level optimization. At the algorithmic level, ALISA employs a Sparse Window Attention mechanism to identify and prioritize crucial tokens for attention computation, while at the system level, it implements a three-phase token-level dynamic scheduler to manage KV tensor allocation and balance caching and recomputation.

\section{Conclusion and Future Work \label{sec:conclusion}}

In this work, we introduce \textbf{BaKlaVa}, a simple yet effective approach to optimize LLM inference through intelligent KV-cache memory allocation. By leveraging a heuristic-based method to estimate layer and KV-cache importance, BaKlaVa significantly improves memory efficiency while maintaining model performance across a range of compression ratios. Our empirical evaluations demonstrate that BaKlaVa outperforms existing KV-cache allocation strategies, such as uniform allocation (StreamingLLM) and allocating KV-cache memory with layer-wise granularity (SqueezeAttention), particularly in tasks where preserving long-range dependencies is crucial. Notably, BaKlaVa maintains near-baseline performance up to 70\% compression, surpassing alternative methods on long-context datasets by preserving essential information and achieving higher accuracy under high compression across multiple tasks

A key advantage of our method is its ability to adapt to different model architectures by \textit{dynamically}  adjusting memory allocation based on computationally inexpensive heuristics. Unlike prior approaches that apply uniform compression or coarse layer-wise compression, BaKlaVa efficiently distributes KV-cache memory to maximize performance under constrained budgets. These improvements highlight the potential of fine-grained memory allocation in enhancing the efficiency of LLM inference without requiring modifications to model architecture or training procedures.

In future work, our aim is to develop a generalized framework for adaptive KV-cache memory allocation, reducing the need for manual parameter tuning. Additionally, extending BaKlaVa to support additional KV eviction policies and dynamically adjusting memory budgets at runtime could further enhance its applicability to real-world deployments.

\section{Acknowledgments \label{sec:Acknowledgments}}

This research used resources of the Argonne Leadership
Computing Facility, a U.S. Department of Energy (DOE)
Office of Science user facility at Argonne National Laboratory
and is based on research supported by the U.S. DOE Office
of Science-Advanced Scientific Computing Research Program,
under Contract No. DE-AC02-06CH11357.

\printbibliography 

\section{Appendix}
\appendix

\section{Full LongBench Results}

 \begin{figure*}[!htb]
     \centering
     \includegraphics[width=1\linewidth]{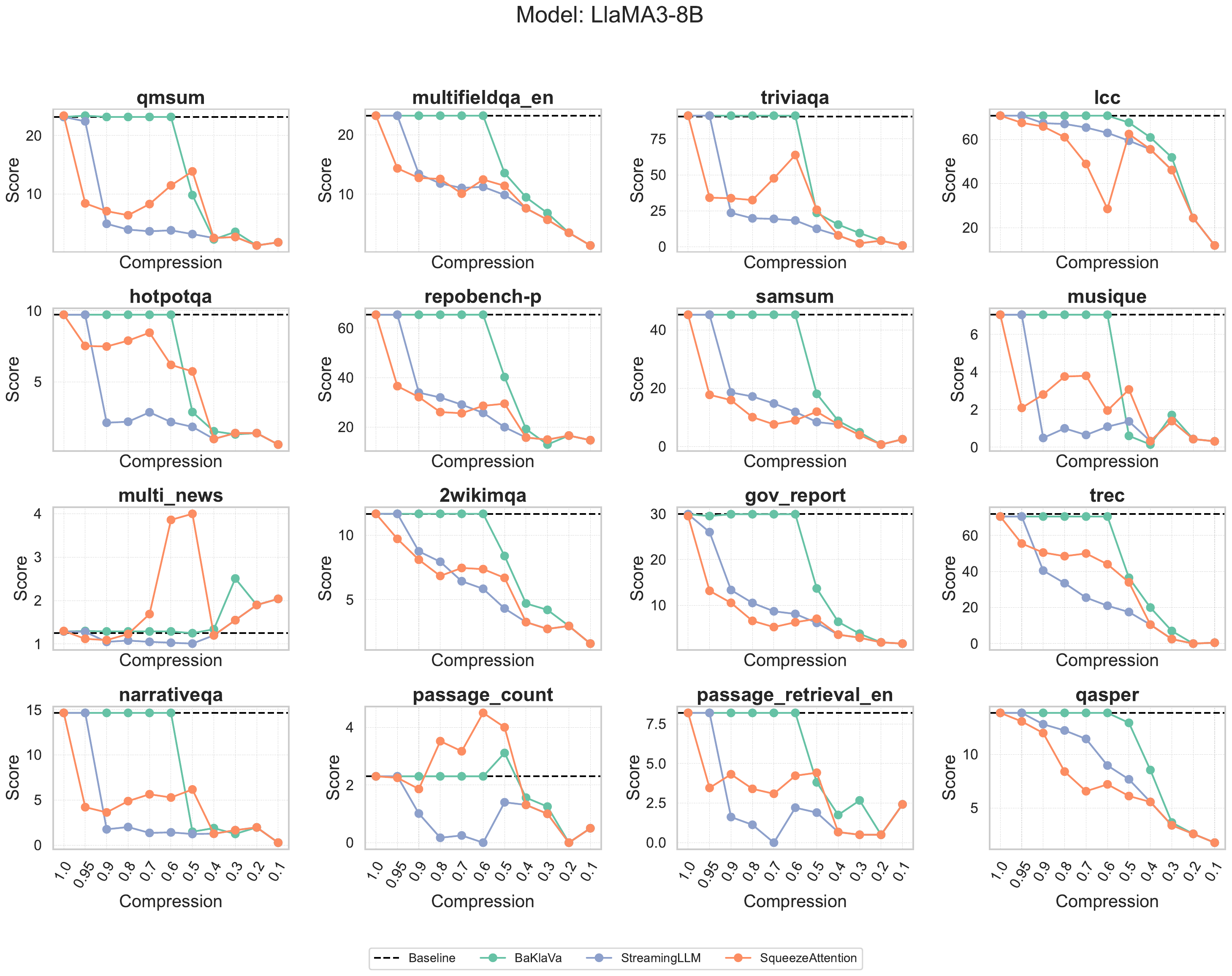}
     \caption{Comparison of BaKlaVa and other cache methods on LlaMA3-8B using LongBench for different compressions.}
     \label{fig:longbench_results_llama}
 \end{figure*}

 \begin{figure*}[!htb]
     \centering
     \includegraphics[width=1\linewidth]{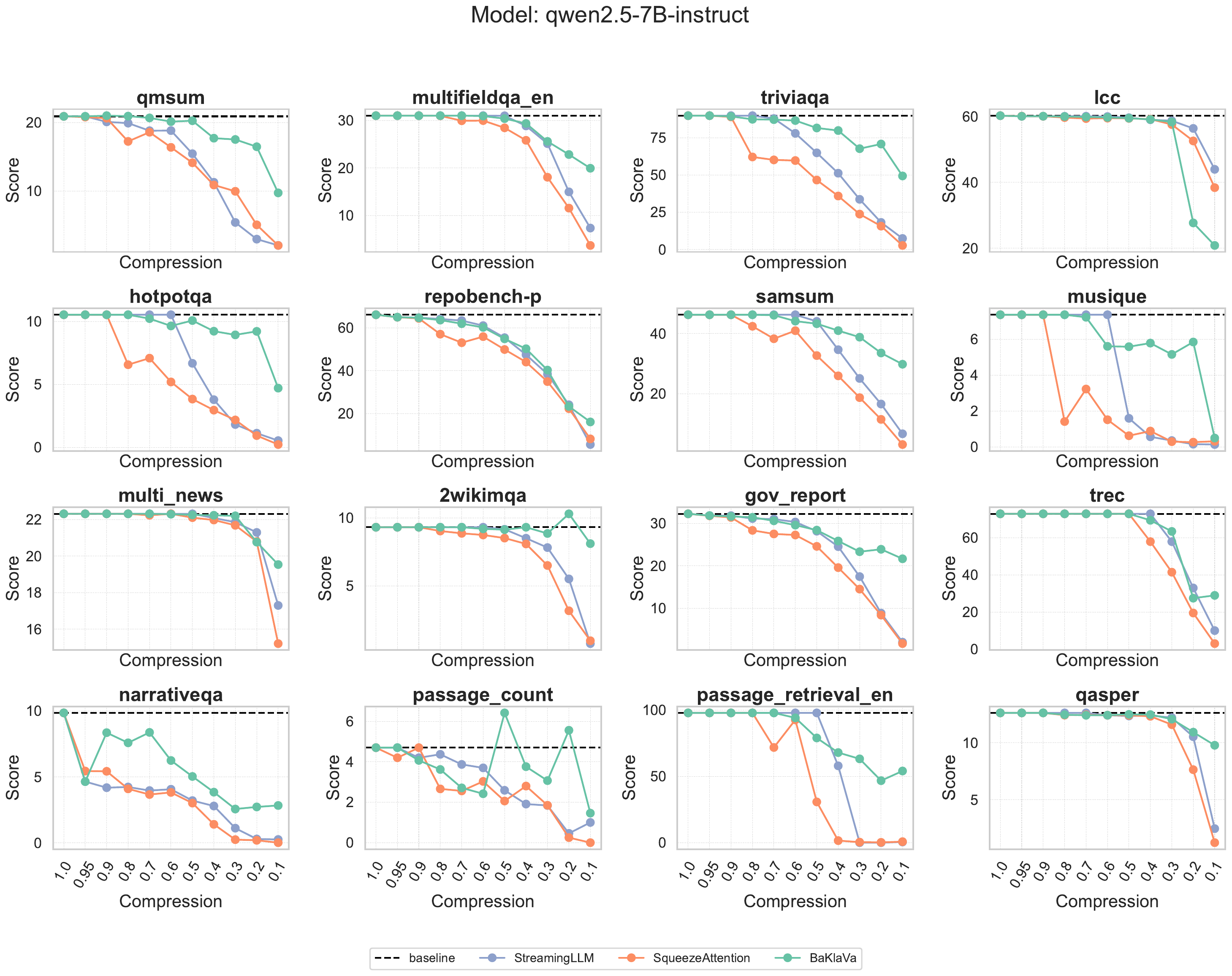}
     \caption{Comparison of BaKlaVa and other cache methods on Qwen2.5-7B using LongBench for different compressions.}
     \label{fig:longbench_results_qwen}
 \end{figure*}

\section{Parameter Search Results}
\begin{figure*}[!htb]
    \centering
    \includegraphics[width=1\linewidth]{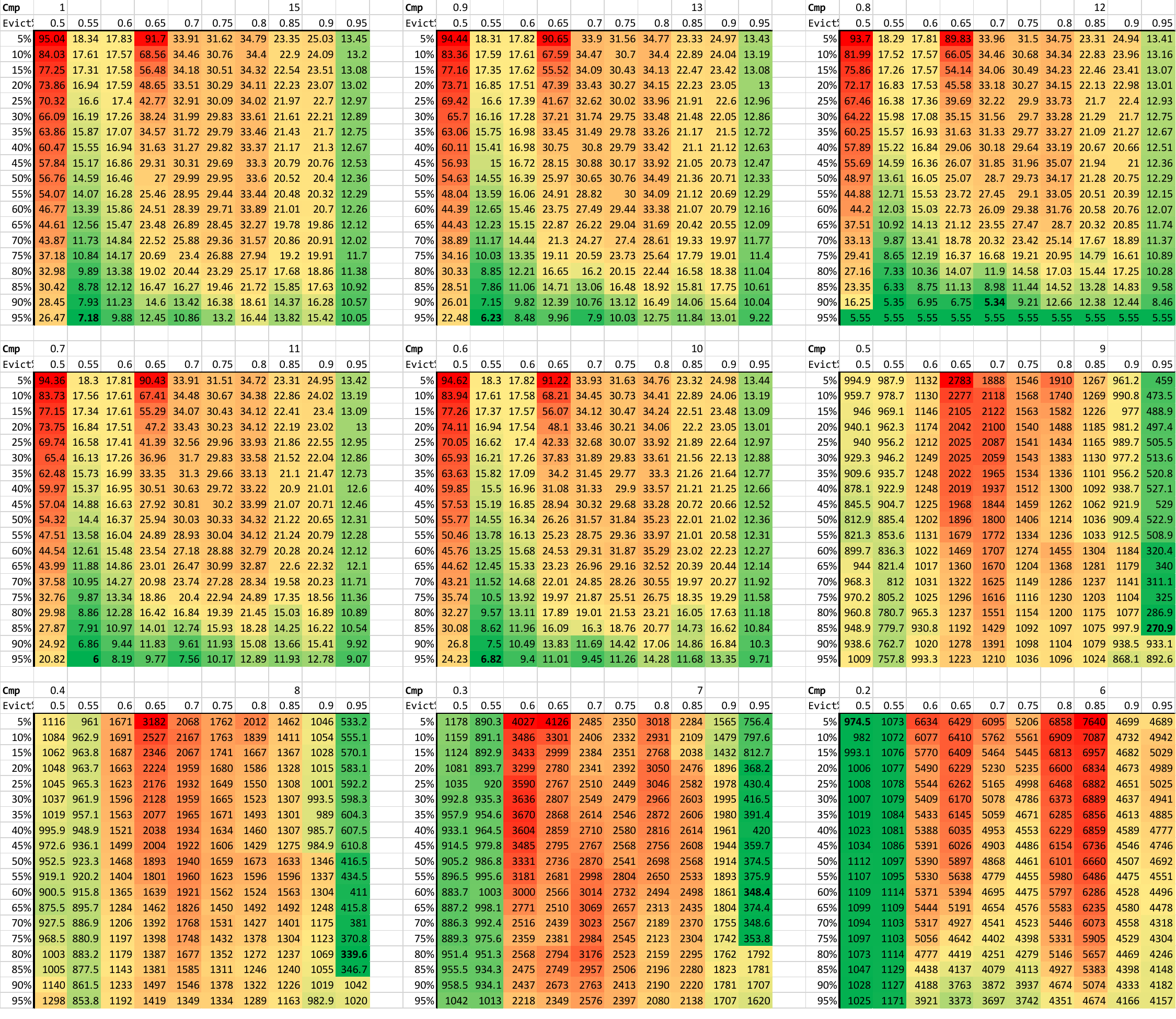}
    \caption{Parameter search results for BaKlaVa method on LlaMA3-8B model. Each compression ratio from 1.0 to 0.2 is shown as a separate heat-map. The X axis is the similarity threshold to select high-similarity/low-importance KV-caches. The Y axis is the low importance/high similarity KV-cache reduction \%. Green regions indicate optimal (low perplexity) parameter configurations, whereas red regions indicate non-optimal (high perplexity) regions. }
    \label{fig:enter-label}
\end{figure*}

\end{document}